\documentclass[twoside]{article}

\usepackage[table]{xcolor}
\usepackage{ecj}
\usepackage{palatino}
\usepackage{natbib}
\usepackage[english]{babel}
\usepackage[utf8]{inputenc}
\usepackage{amsmath}
\usepackage{graphicx}
\usepackage[colorinlistoftodos]{todonotes}
\usepackage{hyphenat}
\usepackage[ruled]{algorithm2e}

\usepackage{rotating}
\usepackage{booktabs}
\usepackage{multirow}
\usepackage{adjustbox}
\usepackage{array}

\ecjHeader{x}{x}{xxx-xxx}{2018}{Embodied Evolution in Collective Robotics -- A Review}{N. Bredeche, E. Haasdijk, A. Prieto}
\title{\bf Embodied Evolution in Collective Robotics:\\A Review}

\author{\name{\bf N. Bredeche} \hfill \addr{nicolas.bredeche@sorbonne-universite.fr}\\ 
        \addr{Sorbonne Universit\'{e}, CNRS, Institute of Intelligent Systems and Robotics (ISIR), F-75005 Paris, France}
\AND
       \name{\bf E. Haasdijk} \hfill \addr{e.haasdijk@vu.nl}\\
        \addr{Department of Computer Science, Vrije Universiteit, Amsterdam, The Netherlands}
\AND
       \name{\bf A. Prieto} \hfill \addr{abraham.prieto@udc.es}\\
        \addr{Integrated Group for Engineering Research, Universidade da Coruna, Ferrol, Spain}
\\
\addr{\vspace{1em} \textit{All authors contributed equally.}}
}

\date{\today}

\begin{document}
\textbf{}\maketitle

\begin{abstract}

This paper provides an overview of evolutionary robotics techniques applied to on-line distributed evolution for robot collectives~--~namely, embodied evolution. 
It provides a definition of embodied evolution as well as a thorough description of the underlying concepts and mechanisms. 
The paper also presents a comprehensive summary of research published in the field since its inception around the year 2000, providing various perspectives to identify the major trends. In particular, we identify a shift from considering embodied evolution as a parallel search method within small robot collectives (fewer than 10 robots) to embodied evolution as an on-line distributed learning method for designing collective behaviours in swarm-like collectives. The paper concludes with a discussion of applications and open questions, providing a milestone for past and an inspiration for future research.
\end{abstract}

\section*{Important notice}

\textit{This paper is a pre-print. The final version has been accepted for publication in Frontiers in Robotics and AI (2018). Please read and use the following updated open-access version for citation:}

\begin{itemize}
\item \textbf{Bredeche N, Haasdijk E and Prieto A (2018) Embodied Evolution in Collective Robotics: A Review. Front. Robot. AI 5:12. doi: 10.3389/frobt.2018.00012}
\end{itemize}
\section{Introduction}
This paper provides an overview of evolutionary robotics research where evolution takes place in a population of robots in a continuous manner.
\citet{Ficici1999} coined the phrase {\em embodied evolution} for evolutionary processes that are distributed over the robots in the population to allow them to adapt autonomously and continuously.
As robotics technology becomes simultaneously more capable and economically viable, individual robots operated at large expense by teams of experts are increasingly supplemented by collectives of robots used cooperatively under minimal human supervision \citep{Bellingham2007a}, and embodied evolution can play a crucial role in enabling autonomous on-line adaptivity in such robot collectives.

The vision behind embodied evolution is one of collectives of truly autonomous robots that can adapt their behaviour to suit varying tasks and circumstances. 
Autonomy occurs at two levels: not only do the robots perform their tasks without external control, they also assess and adapt --through evolution-- their behaviour without referral to external oversight and so learn autonomously.
This adaptive capability allows robots to be deployed in situations that cannot be accurately modelled a priori. This may be because the environment or user requirements are not fully known or it may be due to the complexity of the interactions among the robots as well as with their environment effectively rendering the scenario unpredictable. 
\textbf{}Also, on\hyp{}board adaptivity intrinsically avoids the reality gap \citep{Jakobi1995} that results from inaccurate modelling of robots or their environment when developing controllers before deployment because controllers continue to develop \textit{after} deployment. 
A final benefit is that embodied evolution can be seen as parallelizing the evolutionary process because it distributes the evaluations over multiple robots. \citet{Alba2002} has shown that such parallelism can provide substantial benefits, including superlinear speed-ups. In the case of robots this has the added benefit of reducing the amount of time spent executing poor controllers per robot, reducing wear and tear.

Embodied evolution's on\hyp{}line nature contrasts with `traditional' evolutionary robotics research. Traditional evolutionary robotics employs evolution in the classical sequential centralised optimisation paradigm: parent and survivor selection are centralised and consider the entire population. The `robotics' part entails a series of robotic trials (simulated or not) in an evolution-based search for optimal robot controllers~\citep{Nolfi2000,Bongard2013,Doncieux2015}. In terms of task performance, embodied evolution has been shown to outperform alternative evolutionary robotic techniques in some setups such as surveillance and self-localisation with flying UAVs \citep{Schut2009a,Prieto2016}, especially regarding convergence speed.

To provide a basis for a clear discussion, we define embodied evolution as a paradigm where evolution is implemented in multi-robotic (two or more robots) system. Two robots are already considered a multi-robotic system since it is still possible to distribute an algorithm among them. These systems exhibit the following features:

\begin{description}
\item[Decentralised] There is no central authority that selects parents to produce offspring or individuals to be replaced. Instead, robots assess their performance, exchange and select genetic material autonomously on the basis of locally available information.
\item[On\hyp{}line] Robot controllers change on the fly, as the robots go about their proper actions: evolution occurs during the operational lifetime of the robots and in the robots' task environment. The process continues after the robots have been deployed. 
\item[Parallel] Whether they collaborate in their tasks or not, the population consists of multiple robots that perform their actions and evolve concurrently, in the same environment, interacting frequently to exchange genetic material.
\end{description}

The decentralised nature of communicating genetic material implies that selection is executed locally, usually involving only a part of the whole population \citep{Eiben2007Exploring-selec}, and that it must be performed by the robots themselves. 
This adds a third opportunity for selection in addition to parent and survivor selection as defined for classical evolutionary computing.
Thus, embodied evolution extends the collection of operators that define an evolutionary algorithm (i.e., evaluation, selection, variation and replacement~\citep{eiben2008book}) with \textit{mating} as a key evolutionary operator:

\begin{description}
 \item[Mating] An action where two (or more) robots decide to send and/or receive genetic material, whether this material will or will not be used for generating new offspring. 
 When and how this happens depends on pre\hyp{}defined heuristics, but also on evolved behaviour, the latter determining to a large extent whether robots ever meet to have the opportunity to exchange genetic material.
\end{description}

In the last $20$ years, on\hyp{}line evolutionary robotics in general, and embodied evolution in particular have matured as research fields. 
This is evidenced by the growing number of relevant publications in respected evolutionary computing venues such as in conferences (e.g. ACM GECCO, ALIFE, ECAL and EvoApplications), journals (e.g. Evolutionary Intelligence's special issue on Evolutionary Robotics \citep{Haasdijk2014a}), workshops (PPSN 2014 ER workshop, GECCO 2015 and 2017 Evolving collective behaviours in robotics workshop) and tutorials (ALIFE 2014, GECCO 2015 and 2017, ECAL 2015, PPSN 2016, ICDL-EPIROB 2016).
A Google Scholar search of publications citing the seminal embodied evolution paper by \cite{Watson2002a} illustrates this growing trend. Since 2009, the paper has attracted substantial interest, more than doubling the yearly number of citations since 2008 (approx. 20 citations per year since then). \footnote{See \texttt{https://plot.ly/\textasciitilde evertwh/17/} for more details and the underlying data.}

To date, however, a clear definition of what embodied evolution is (and what it is not) and an overview of the state of the art in this area are not available.
This paper provides a definition of the embodied evolution paradigm and relates it to other evolutionary and swarm robotics research (sections \ref{sec:context} and \ref{sec:algorithmic description}). 
We identify and review relevant research, highlighting many design choices and issues that are particular to the embodied evolution paradigm (sections \ref{sec:state-of-the-art} and \ref{sec:discussionposttable}).
Together this provides a thorough overview of the relevant state-of-the-art and a starting point for researchers interested in evolutionary methods for collective autonomous adaptation.
Section \ref{sec:discussion} identifies open issues and research in other fields that may provide solutions, suggests directions for future work and discusses potential applications.

\section{Context}
\label{sec:context}

Embodied Evolution considers collectives of robots that adapt on-line.
This section positions embodied evolution vis \`a vis other methods for developing  controllers for robot collectives and for achieving on-line adaptation.

\subsection{Off-line Design of Behaviours in Collective Robotics}

Decentralised decision-making is a central theme in collective robotics research: when the robot collective cannot be centrally controlled, the individual robots' behaviour must be carefully designed so that global coordination occurs through local interactions.

Seminal works from the 1990s such as \citeauthor{Mataric1994}'s Nerd Herd (\citeyear{Mataric1994}) addressed this problem by hand-crafting behaviour-based control architectures.
Manually designing robot behaviours has since been extended with elaborate methodologies and architectures for multi-robot control (see \cite{Parker2008a} for a review) and with a plethora of bio-inspired control rules for swarm-like collective robotics (see \cite{Nouyan2009,Rubenstein2014} for recent examples involving real robots, and \cite{Beni2005,Brambilla2012,Bayindir2016} for discussions and recent reviews).

Automated design methods have been explored with the hope of tackling problems of greater complexity. Early examples of this approach were applied to the robocup challenge for learning coordination strategies in a well-defined setting. 
See \cite{Stone1998} for an early review and \cite{Stone2005a} and \cite{Barrett2016} for more recent work in this vein.
However, \citet{Bernstein2002} demonstrated that solving even the simplest multi-agent learning problem is intractable in polynomial time (actually, it is NEXP-complete), so obtaining an optimal solution in reasonable time is currently infeasible. 
Recent works in reinforcement learning have developed theoretical tools to break down complexity by operating a move from considering many agents to a collection of single agents, each of which being optimised separately~\citep{Dibangoye2014}, leading to theoretically well-founded contributions, but with limited practical validation involving very few robots and simple tasks~\citep{amato2015}. 

Lacking theoretical foundations, but instead based on experimental validation, swarm robotics controllers have been developed with black-box optimisation methods ranging from brute-force optimisation using a simplified (hence tractable) representation of a problem \citep{Werfel2014a} and evolutionary robotics \citep{Trianni2008,Hauert2008,gauci2013evolving,Silva2016}. 

The methods vary, but all the approaches described here (including `standard' evolutionary robotics) share a common goal: to design or optimise a set of control rules for autonomous robots that are part of a collective \textit{before} the actual deployment of the robots. 
The particular challenge in this kind of work is to design individual behaviours that lead to some required global (`emergent') behaviour without the need for central oversight. 


\subsection{Lifelong Learning in Evolutionary Robotics} 

It has long been argued that deploying robots in the real world may benefit from continuing to acquire new capabilities \textit{after} initial deployment~\citep{Thrun1995,Nelson2006}, especially if the environment is not known beforehand. 
Therefore, the question we are concerned with in this paper is \textit{how to endow a collective robotics system with the capability to perform lifelong learning}. 
Evolutionary robotics research into this question typically focuses on individual autonomous robots. 
Early works in evolutionary robotics that considered lifelong learning 
explored learning mechanisms to cope with minor environmental changes (see the classic book by \cite{Nolfi2000} as well as \cite{Urzelai2001b} and \cite{Tonelli2013c} for examples, and \cite{Mouret2015} for a nomenclature).
More recently, \cite{Bongard2006b} and \cite{Cully2014b} addressed \textit{resilience} 
by introducing fast on-line re-optimisation to recover from hardware damage. 

\cite{BredecheOn-line-On-boar}, \cite{Christensen2010} and \cite{Silva2012} are some examples of on\hyp{}line versions of evolutionary robotics algorithms that target the fully autonomous acquisition of behaviour to achieve some pre-defined task in individual robots.
Targeting agents in a video game rather than robots, \cite{Stanley2005} tackled the on-line evolution of controllers in a multi-agent system. 
Because the agents were virtual, the researchers could control some aspects of the evaluation conditions (e.g.,  restarting the evaluation of agents from the same initial position). This kind of control is typically not feasible in autonomously deployed robotic systems.

Embodied evolution builds on evolutionary robotics to implement lifelong learning in robot {\em collectives}. 
Its clear link with traditional evolutionary robotics is exemplified by work like that by \citet{Usui2003Situated-and-Em}, where a traditional evolutionary algorithm is encapsulated on each robot. Individual controllers are evaluated sequentially in a standard time sharing set-up, and the robots implement a communication scheme that resembles an island model to exchange genomes from one robot to another. It is this communication that makes this an instance of embodied evolution.

 
\section{Algorithmic description}
\label{sec:algorithmic description}

This section presents a formal description of the embodied evolution paradigm by means of generic pseudo-code and a discussion about its operation from a more conceptual perspective.

The pseudo-code in algorithm \ref{alg:embodied evolution} provides an idealised description of a robot's control loop as it pertains to embodied evolution. Each robot runs its own instance of the algorithm, and the evolutionary process emerges from the interaction between the robots. In embodied evolution, there is no entity outside the robots that oversees the evolutionary process and there is typically no synchronisation between the robots: the replacement of genomes is asynchronous and autonomous.

\begin{algorithm*}[t]
\BlankLine
initialise robot\;
\BlankLine
\For(){ever}
{
    Sense - Act cycle (depends on robotic paradigm)\;
    \BlankLine									
	perf $\leftarrow$ calculate performance\;
	\BlankLine									
	\If(\tcp*[f]{E.g., is another robot nearby?}){ mating? } 	
	{
		transmit my genome\tcp*{and optional further information}
		$g$ $\leftarrow$ receive mate's genome\;
		store($g$)\;
	}
	\BlankLine									
	\If(\tcp*[f]{E.g., time or virtual energy runs out}){replacement? }     
	{
		parents $\leftarrow$ select parents\;
		offspring $\leftarrow$ variation(parents)\;        
        activate(offspring)\tcp{Time-sharing: control is handed over to the new candidate controller}
	}
	\BlankLine									
}
\caption{An individual robot's control loop for embodied evolution.}
\label{alg:embodied evolution}
\end{algorithm*}

Some steps in this generic control loop can be implicit or entwined in particular implementations. For instance, robots may continually broadcast genetic material over short range so that other robots that come within this range receive it automatically. In such a case, the $mating$ operation is implicitly defined by the selected broadcast range.
Similarly, genetic material may be incorporated into the currently active genome as it is received, merging the mating and replacement operations.
Implicitly defined or otherwise, the steps in this algorithm are, with the possible exception of performance calculation, necessary components of any embodied evolution implementation.

The following list describes and discusses the steps in the algorithm in detail.

\begin{description}

\item[Initialisation] The robot controllers are typically initialised randomly, but it is possible that the initial controllers are developed off-line, be it through evolution or hand-crafted (e.g.: see the work from \citet{Hettiarachchi2006}). 

\item[Sense--act cycle] This represents `regular', i.e., not related to the evolutionary process, robot control. The details of the sense-act cycle depend on the robotic paradigm that governs robot behaviour; this may include planning, subsumption or other paradigms. This may also be implemented as a separate parallel process.

\item[Calculate performance] If the evolutionary process defines an objective function, the robots monitor their own performance. 
This may involve measurements of quantities such as speed, number of collisions or amount of collected resources. 
Whatever their nature, these measurements are then used to evaluate and compare genomes (as fitness values  in  evolutionary computation).
The possible discrepancy between the individual's objective function and the population welfare will be discussed further in Section~\ref{sec:openissues}.

\item[Mating] This is the essential step in the evolutionary process where robots exchange genetic material. The choice to mate with another robot may be purely based on environmental contingencies (e.g., when robots mate whenever they are within communication range), but other considerations may also play a part (e.g., performance, genotypic similarity, etc.). 
The pseudo-code describes a symmetric exchange of genomes (with both a transmit and a receive operation), but this may be asymmetrical for particular implementations.
In implementations such as that of \cite{Schwarzer2011} or \cite{Haasdijk2014}, for instance, robots suspend normal operation to collect genetic material from other, active robots.
Mating typically results in a pool of candidate parents that are considered in the parent selection process.

\item[Replacement] The currently active genome is replaced by a new individual (the offspring), implying the removal of the current genome. This event can be triggered by a robot's internal conditions (e.g., running out of time or virtual energy, reaching a given performance level) or through interactions with other robots (e.g., receiving promising genetic material~\citep{Watson2002a}).

\item[Parent selection] 
This is the process that selects which genetic information will be used for the creation of new offspring from the received genetic information through mating events.
When an objective is defined, the performance of the received genome is usually the basis for selection, just as in regular evolutionary computing. In other cases, the selection among received genomes can be random or depend on non-performance related heuristics (e.g., random, genotypic proximity, etc.). In the absence of objective-driven selection pressure, individuals are still competing with respect to their ability to spread their own genome within the population, though that cannot be explicitly capture during parent selection. This will be further discussed in section~\ref{subsec:selectionpressure}.

\item[Variation] A new genome is created by applying the variation operators (mutation and crossover) on the selected parent genome(s). This is subsequently activated to replace the current controller.

\end{description}

\begin{figure}[!ht]
\begin{center}
\includegraphics[width=\linewidth]{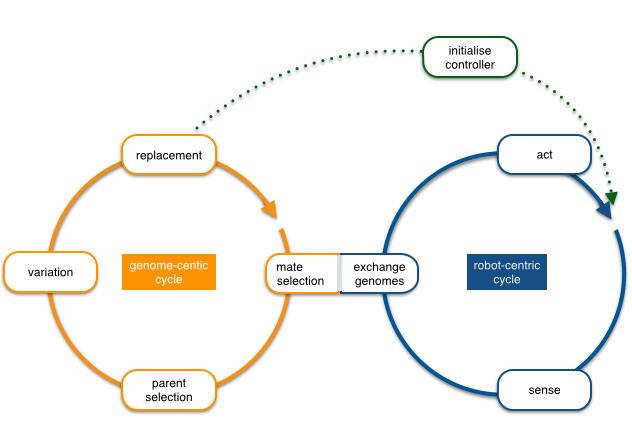}

 \caption{The overlapping robot-centric and genome-centric cycles in embodied evolution. The robot-centric cycle uses a single
\textit{active} genome that determines the current robot behaviour (sense-act loop), the genome-centric cycle manages an internal \textit{reservoir} of genomes received from other robots or built locally (parent selection / variation), out of which the next active genome will be selected eventually (replacement).}  
 
\label{fig:embodied_evolution_cycle}
\end{center}
\end{figure}

\noindent From a conceptual perspective, embodied evolution can be analysed at two levels which are represented by two as depicted in Fig.\ref{fig:embodied_evolution_cycle}:

\paragraph*{The robot-centric cycle} is depicted on the right in Fig.\ref{fig:embodied_evolution_cycle}. It represents the physical interactions that occur between the robot and its environment, including interactions with other robots and extends this sense-act loop commonly used to describe real-time control systems by accommodating the exchange and activation of genetic material. At this particular point, the genome-centric and robot-centric cycles overlap.
The cycle operates as follows: each robot is associated to an \textit{active} genome, the genome is interpreted into a set of features and control architecture (the phenotype) which produces a behaviour which includes the transmission of its own genome to some other robots. Each robot eventually switches from an active genome to another, depending on a specific event (e.g. minimum energy threshold) or duration (e.g. fixed lifetime), and consequently changes its active genome, probably impacting its behaviour.

\paragraph*{The genome-centric cycle} deals with the events that directly affect the genomes existing in the robot population and therefore also the evolution per se. Again, the mating and the replacement are the events which overlap with the robot-centric cycle.
The operation from the genome cycle perspective is as follows: each robot starts with an initial genome, either initialised randomly or a priori defined. While this genome is \textit{active} it determines the phenotype of the robot, hence its behaviour. Afterwards, when replacement
is triggered, some genomes are selected from the reservoir of genomes previously received according to the parent selection criteria and later combined using the variation operators. This new genome will then become part of the population. In the case of fixed size population algorithms, the replacement
will automatically trigger the removal of the old genome.
In some other cases, however, there is a specific criterion to trigger the removal event producing populations of individuals which change their size along the evolution. 

The two circles connect on two occasions, firstly by the `exchange genomes’ (or mating) process which implies the transmission of genetic material, possibly together with additional information (fitness if available, general performance, genetic affinity, etc.) to modulate future selection. 
Generally, the received information is stored to be used (in full or in part) to replace the active genome in the later parent selection process. Therefore, the event is triggered and modulated by the robot cycle but it impacts on the genomic cycle. Also, the decentralised nature of the paradigm enforces that these transmissions occur locally, either one-to-one or to any robot in a limited range. 
There are several ways in which mate selection can be implemented, for instance, individuals may send and receive genomic information indiscriminately within a certain location range or the frequency of transmission can depend on the task performance.
The second overlap between the two cycles is the activation of new genomic information (replacement). The activation of a genome in the genomic-cycle implies that this new genome will now take control of the robot and therefore changes the response of the robot in the scenario (in evolutionary computing terms this event will mark the start of a new individual evaluation). This aspect is what creates the on\hyp{}line character of the algorithm which, together with the locality constraints, implies that the process is also asynchronous. 

\paragraph*{}This conceptual representation matches what has been defined as \textit{distributed} embodied evolution by~\cite{Eiben2010Embodied-On-lin}. The authors proposed a taxonomy for on\hyp{}line evolution that differentiates between encapsulated, distributed and hybrid schemes. Most embodied evolution implementations are distributed, but this schematic representation also covers hybrid implementations. In such cases, the robot locally maintains a population that is augmented through mating (rather like an island model in parallel evolutionary algorithms). It should be noted that encapsulated implementations (where each robot runs independently of the others) are not considered in this overview.


\section{Embodied Evolution: The State of the Art}
\label{sec:state-of-the-art}



\newcolumntype{R}[2]{%
    >{\adjustbox{angle=#1,lap=\width-(#2)}\bgroup}%
    c%
    <{\egroup}%
}
\newcommand*\rot{\multicolumn{1}{R{-90}{.01cm}}}
\newcommand*\Yes{$\bullet$}

\begin{sidewaystable}[t]
\scriptsize
\begin{center}
\begin{tabular}{p{2in}p{.01cm}p{.01cm}p{.01cm}p{.01cm}cccccp{1.2in}lcccccccccc}
\toprule
\multicolumn{5}{r}{Implementation}
&\multicolumn{5}{|c}{Robot Behaviour}
&\multicolumn{2}{|c}{Experimental Settings}
&\multicolumn{3}{|c}{Mating Conditions}
&\multicolumn{3}{|c}{Selection Scheme}
&\multicolumn{4}{|c}{Replacement Scheme}
\\ \midrule

&\rot{distributed}&\rot{hybrid}
&\rot{simulation}&\rot{real robots}
&\rot{monomorphic}&\rot{polymorphic}
&\rot{individual behaviour}&\rot{cooperation}&\rot{division of labour}
&\rot{task}~
&\rot{nr of robots}
&\rot{panmictic}&\rot{proximity}&\rot{other}
&\rot{performance}&\rot{random}&\rot{genotypic distance}
&\rot{fixed lifetime}&\rot{variable lifetime}&\rot{event-based}&\rot{limited lifetime}
\\

\rowcolor{black!15}
\cite{Ficici1999,Watson2002a} 
&\Yes 	
&~ 		
&~		
&\Yes	
&\Yes	
&~		
&\Yes	
&~		
&~		
&phototaxis 
&8 		
&~		
&\Yes	
&~		
&\Yes	
&~		
&~		
&		
&\Yes	
&~		
&~		
\\

\cite{Simoes2001}	
&\Yes&~ 
&~&\Yes 
&\Yes&~&\Yes&&~
&obstacle avoidance
&6 
&\Yes&~&~ 
&\Yes&~&~ 
&\Yes&~&~&~ 
\\

\rowcolor{black!15}
\cite{Usui2003Situated-and-Em}	
&	 	
&\Yes	
&~		
&\Yes	
&\Yes	
&~		
&\Yes	
&~		
&~		
&obstacle avoidance 
&6 		
&~		
&\Yes	
&~		
&\Yes	
&~		
&~		
&\Yes	
&		
&~		
&~		
\\

\cite{Bianco2004}
&\Yes 	
&		
&\Yes	
&		
&\Yes	
&~		
&\Yes	
&~		
&~		
&self assembly 
&64		
&~		
&\Yes	
&~		
&~		
&\Yes	
&		
&		
&		
&\Yes	
&~		
\\

\rowcolor{black!15}
\cite{Hettiarachchi2006,Hettiarachchi2009a}	
&\Yes		
&		
&\Yes	
&		
&\Yes 	
&~		
&\Yes	
&~		
&~		
&navigation with obstacle avoidance 
&60	
&		
&\Yes	
&~		
&\Yes	
&~		
&~		
&		
&		
&\Yes		
&~		
\\

\cite{Wischmann2007Embodied-evolut}	
&\Yes&~ 
&\Yes&~ 
&\Yes&~&\Yes&~&~
&foraging$^1$
&3
&~&\Yes&~
&\Yes&~&~ 
&~&\Yes&~&~ 
\\

\rowcolor{black!15}
\cite{Perez2008Embodied-Evolut}	
& 		
&\Yes	
&\Yes	
&		
&\Yes 	
&~		
&\Yes	
&~		
&~		
&obstacle avoidance 
&5		
&\Yes	
&		
&~		
&\Yes	
&~		
&~		
&\Yes	
&		
&		
&~		
\\

\cite{Konig2009a,Konig2009b}
&\Yes		
&		
&\Yes	
&		
&\Yes 	
&~		
&\Yes	
&~		
&~		
&obstacle avoidance with gate passing 
&26;30	
&		
&\Yes	
&~		
&\Yes	
&~		
&~		
&\Yes	
&		
&		
&~		
\\

\rowcolor{black!15}
\cite{Pugh2009} 
&		
&\Yes	
&\Yes	
&\Yes	
&\Yes 	
&~		
&\Yes	
&~		
&~		
&obstacle avoidance
&1--10	
&\Yes		
&\Yes	
&~		
&\Yes	
&~		
&~		
&\Yes	
&		
&		
&~		
\\

\cite{Prieto2009,Trueba2011,Trueba2012a}	
&\Yes		
&	
&\Yes	
&	
& 	
&\Yes		
&	
&\Yes		
&\Yes		
&surveillance, foraging, construction
&20	
&		
&\Yes	
&~		
&\Yes	
&~		
&~		
&		
&		
&\Yes		
&\Yes		
\\

\rowcolor{black!15}
\cite{bredeche2010ppsn,Bredeche2012,bredeche2014alife}	
&\Yes		
&	
&\Yes	
&\Yes	
&\Yes 	
&		
&	
&\Yes		
&~		
&none
&20;4000	
&		
&\Yes	
&~		
&	
&\Yes		
&~		
&		
&		
&		
&\Yes		
\\

\cite{Prieto2010}	
&\Yes		
&	
&	
&\Yes	
& 	
&\Yes	
&	
&\Yes		
&\Yes		
&surveillance
&8	
&		
&\Yes	
&~		
&\Yes	
&		
&~		
&		
&		
&\Yes		
&\Yes		
\\

\rowcolor{black!15}
\cite{Schwarzer2010Artificial-Sexu} 
&\Yes		
&	
&\Yes	
&\Yes	
& 	
&\Yes	
&	
&\Yes		
&~		
&none
&upto 40	
&		
&\Yes	
&~		
&	
&\Yes		
&		
&		
&\Yes		
&		
&		
\\

\cite{Schwarzer2011}
&\Yes		
&	
&\Yes	
&	
&\Yes 	
&	
&\Yes	
&		
&~		
&phototaxis
&4	
&		
&\Yes	
&~		
&\Yes	
&		
&		
&\Yes	
&		
&		
&		
\\

\rowcolor{black!15}
\cite{Montanier2011,montanier2013}
&\Yes		
&	
&\Yes	
&	
&\Yes 	
&	
&	
&\Yes		
&~		
&none
&100	
&		
&\Yes	
&~		
&	
&\Yes		
&\Yes		
&		
&		
&		
&\Yes		
\\

\cite{Huijsman2011An-On-line-On-b} 
&\Yes		
&\Yes	
&\Yes	
&	
&\Yes 	
&	
&\Yes	
&		
&~		
&obstacle avoidance
&4--400	
&\Yes		
&	
&\Yes		
&\Yes	
&		
&		
&\Yes		
&		
&		
&		
\\

\rowcolor{black!15}
\cite{Karafotias2011An-Algorithm-fo}
&		
&\Yes	
&\Yes	
&	
&\Yes 	
&	
&\Yes	
&		
&\Yes	
&obstacle avoidance, phototaxis and patrolling
&10	
&		
&	
&\Yes		
&\Yes	
&		
&		
&\Yes		
&		
&		
&		
\\

\cite{Silva2012,Silva2013a,Silva2015,Silva2017}	
&		
&\Yes	
&\Yes	
&\Yes	
& 	
&\Yes	
&\Yes	
&\Yes		
&~		
&navigation, aggregation, surveillance, phototaxis
&2--20	
&		
&\Yes	
&~		
&\Yes	
&		
&		
&		
&		
&\Yes		
&		
\\

\rowcolor{black!15}
\cite{weel2012organisms,weel2012online}
&		
&\Yes	
&\Yes	
&	
&\Yes 	
&	
&	
&\Yes		
&~		
&foraging
&10;50	
&		
&\Yes	
&~		
&\Yes	
&		
&		
&\Yes		
&		
&		
&		
\\

\cite{Garcia-Sanchez2012Testing-Diversi}
&		
&\Yes	
&\Yes	
&	
&\Yes 	
&	
&\Yes	
&		
&~		
&obstacle avoidance
&4--36	
&		
&\Yes	
&\Yes	
&\Yes	
&		
&		
&\Yes		
&		
&		
&		
\\

\rowcolor{black!15}
\cite{Noskov2013a,Haasdijk2013,Haasdijk2013b,Haasdijk2014,Haasdijk2016,Kemeling2017,Bangel2017}
&\Yes		
&	
&\Yes	
&	
& 	
&\Yes	
&\Yes	
&		
&\Yes	
&foraging
&100	
&		
&\Yes	
&~		
&\Yes	
&		
&		
&\Yes		
&		
&		
&		
\\

\cite{Trueba2013682,Trueba2017}
&\Yes	
&	
&\Yes	
&\Yes	
& 	
&\Yes	
&	
&\Yes		
&\Yes~		
&synthetic mapping, gathering, self-location
&40;20;9	
&\Yes		
&\Yes	
&~		
&\Yes	
&\Yes		
&		
&		
&		
&\Yes		
&\Yes		
\\

\rowcolor{black!15}
\cite{ODowd2014a}
&		
&\Yes	
&	
&\Yes	
& \Yes	
&	
&\Yes	
&		
&~		
&foraging
&10	
&		
&\Yes	
&~		
&\Yes	
&		
&		
&\Yes		
&		
&		
&		
\\

\cite{FernandezPerez2014}
&\Yes		
&	
&\Yes	
&	
&\Yes	
&	
&	
&		
&\Yes		
&foraging	
&50	
&		
&\Yes	
&~		
&\Yes	
&\Yes		
&		
&\Yes		
&		
&		
&		
\\

\rowcolor{black!15}
\cite{FernandezPerez2015}
&\Yes		
&	
&\Yes	
&	
&\Yes 	
&	
&	
&\Yes		
&~		
&foraging
&100	
&		
&\Yes	
&~		
&\Yes	
&		
&		
&		
&		
&\Yes	
&		
\\

\cite{Hart2015,Steyven2016}
&\Yes		
&	
&\Yes	
&	
&\Yes 	
&	
&	
&\Yes		
&~		
&foraging
&100	
&		
&\Yes	
&~		
&\Yes	
&		
&		
&		
&		
&		
&\Yes	
\\

\rowcolor{black!15}
\cite{Heinerman2015-3foldadaptivity,Heinerman2015-3foldthymio}
&		
&\Yes	
&\Yes	
&\Yes	
&\Yes 	
&	
&\Yes	
&		
&~		
&obstacle avoidance
&6	
&\Yes		
&	
&~		
&\Yes	
&		
&		
&\Yes		
&		
&		
&		
\\

\cite{Montanier2016,Bredeche2017}
&\Yes		
&	
&\Yes	
&	
&	
&\Yes 	
&	
&		
&\Yes		
&foraging
&100;500	
&		
&\Yes	
&~		
&\Yes	
&\Yes	
&		
&		
&		
&		
&\Yes		
\\
\rowcolor{black!15}
\cite{Perez2017}
&\Yes		
&	
&\Yes	
&	
&	
&\Yes 	
&	
&\Yes		
&		
&foraging
&200	
&		
&\Yes	
&~		
&\Yes	
&	
&		
&\Yes		
&		
&		
&		
\\
\bottomrule
\multicolumn{8}{l}{
$^1$ as a proxy for predator avoidance}\\
\end{tabular}
\end{center}
\caption{Overview of Embodied Evolution research}
\label{tab:overview}
\end{sidewaystable}


\setlength{\extrarowheight}{0.1cm}
\begin{table}[ht]
\centering
\begin{tabular}{lp{10cm}}
\toprule
\textbf{Field} & 
\textbf{Comment} \\
\midrule
Implementation & \textbf{Distributed} implementations have one genome for each robot, and an offspring is created only as the result of a mating event or by mutating the current genome. \textbf{Hybrid} implementations have multiple genomes per robot, and offspring can be created from this internal pool as well as from genomes `imported' through mating events. As stated earlier, the encapsulated scheme is not considered embodied evolution as there is no exchange of genomes between robots in this case.\\
 & The experiments can use \textbf{real robots} or \textbf{simulation}.\\
Robot behaviour & 
A \textbf{monomorphic} population contains individuals with similar genotypes (with variations due to mutation). A \textbf{polymorphic} population is divided into two (or more) subgroups of genetically similar individuals, and different genotypic signatures from one group to the other, e.g., to achieve specialisation.\\
 & We distinguish between experiments that target efficient \textbf{individual behaviour} vs. collective behaviours (i.e., social behaviours, incl. cooperation) \\
Experimental settings & Identifies the \textbf{task(s)} considered in the experiment, e.g., obstacle avoidance, foraging, ... \textbf{None} indicates that there is no user-defined task and that consequently, selection pressure results from the environment only. 
The \textbf{number of robots} used is also included. $n_1-n_2$ indicates the interval for one experiment and $n1;n2$ gives numbers for two experiments.\\
Mating conditions & Mating can be based on \textbf{proximity}: two robots can mate whenever they are physically close to each other (e.g., in infrared communication range). In \textbf{panmictic} systems, robots can mate with all other robots, regardless of their location. \textbf{Other} comprises of systems where robots maintain an explicit list of potential mates (a social network) which may be maintained through gossiping.  \\
Selection scheme & Parents are selected from the received and internal genomes on the basis of their \textbf{performance} if a task is defined. \textbf{Random} parent selection implies only environment-driven selection. Currently, the only examples of \textbf{other} selection schemes use genotypic distance, but this category also covers metrics such as novelty. \\
Replacement scheme & Genomes can have a \textbf{fixed lifetime}, \textbf{variable lifetime} or \textbf{limited lifetime} (similar to variable lifetime, but with an upper bound). \textbf{Event\hyp{}based} replacement schemes do not depend on time, but on events such as reception of genetic material (e.g. in the microbial GA used by \citet{Watson2002a}).  \\
\bottomrule
\end{tabular}
\caption{Glossary}
\label{tab:glossary}
\end{table}


In this Section, we identify the major research topics from the works published since the inception of the domain, all summarized in table~\ref{tab:overview}.
Table \ref{tab:overview} provides an overview of published research on embodied evolution with robot collectives. Each entry describes a contribution, which may cover several papers. The entries are described in terms of their implementation details, the robot behaviour, experimental settings, mating conditions, selection and replacement schemes. The glossary in table \ref{tab:glossary} explains these features in more detail.

We firstly distinguish between works that consider embodied evolution as a parallel search method for optimising \textit{individual} behaviours and works where embodied evolution is employed to craft \textit{collective} behaviour in robot populations. The latter trend, where the emphasis is on collective behaviour, has emerged relatively recently and since then has gained importance (32 papers since 2009).

We secondly consider the homogeneity of the evolving population; borrowing definitions from biology, we use the term \textit{monomorphic} (resp. \textit{polymorphic}) for a population containing one (resp. more than one) class of genotype, for instance to achieve specialisation.
A monomorphic population implies that individuals will behave in a similar manner (except for small variations due to minor genetic differences). On the contrary, polymorphic populations host multiple groups of individuals, each group with its particular genotypic signature, possibly displaying a specific behaviour. Research to date shows that cooperation in monomorphic populations can be easily achieved (e.g.~\cite{Prieto2010,Schwarzer2010Artificial-Sexu,Montanier2011,montanier2013,Silva2012}), while polymorphic populations (e.g., displaying genetic-encoded behavioural specialisation) require very specific conditions to evolve (e.g.,~\cite{Trueba2013682,Haasdijk2014,Montanier2016}).

A notable number of contributions employ real robots. Since the first experiments in this field, the intrinsic on\hyp{}line nature of embodied evolution has made such validation comparatively straightforward \citep{Ficici1999,Watson2002a}. 
`Traditional' evolutionary robotics is more concerned with robustness at the level of the evolved \textit{behaviour} (mostly caused from the reality gap that exists between simulation and the real world) than is embodied evolution, which emphasises the design of robust \textit{algorithms}, where transfer between simulation and real world may be less problematic.
In the contributions presented here, simulation is used for extensive analysis that could hardly take place with real robots due to time or economic constraint. 
Still, it is important to note that many researchers who use simulation have also published works with real robots, thus including real world validation in their research methodology.

Since 2010, there has been a number of experiments that employ large ($\geq 100$) numbers of (simulated) robots, shifting towards more swarm-like robotics where evolutionary dynamics can be quite different~\citep{Huijsman2011An-On-line-On-b,bredeche2014alife,Haasdijk2014a}.
Recent works in this vein focus on the nature of selection pressure, emphasising the unique aspect of embodied evolution that selection pressure results from both the environment (which impacts mating) and the task. 
\citet{bredeche2010ppsn,Bredeche2012} showed that environmental pressure alone can drive evolution towards self-sustaining behaviours.
\citet{Haasdijk2014} showed that these selection pressures can to some extent be modulated by tuning the ease with which robots can transmit genomes. \citet{Steyven2016} showed that adjusting the availability
and value of energy resources results in the evolution of a range of different behaviours.
These results emphasise that tailoring the environmental requirements to transmit genomes can profoundly impact the evolutionary dynamics, and that understanding these effects is vital to effectively develop embodied evolution systems.

\section{Issues in Embodied Evolution}
\label{sec:discussionposttable}

What sets embodied evolution apart from classical evolutionary robotics (and, indeed, from most evolutionary computing) is the fact that evolution acts as a force for continuous adaptation, not (just) as an optimiser before deployment.
As a continuous evolutionary process, embodied evolution is similar to some evolutionary systems considered in artificial life research (e.g., \cite{axelrod1984the-evolution-o,Ray1993}, to name a few).
The operations that implement the evolutionary process to adapt the robots' controllers are an integral part of their behaviour in their task environment.
This includes mating behaviour to exchange and select genetic material, assessing one's own and/or each other's task performance (if a task is defined) and applying variation operators such as mutation and recombination.

This raises issues that are particular to embodied evolution. 
The research listed in the previous section has identified and investigated a number of these issues, and the remainder of this section discusses these issues in detail, while section \ref{sec:openissues} discusses issues that so far have not benefited from close attention in embodied evolution research.

\subsection{Local Selection}
In embodied evolution, the evolutionary process is generally implemented through local interactions between the robots, i.e., the mating operation introduced above. This implies the concept of a neighbourhood from which mates are selected. One common way to define neighbourhood is to consider robots within communication range, but it can also be defined in terms of other distance measures such as genotypic or phenotypic distance. 
Mates are selected by sampling from this neighbourhood and a new individual is created by applying variation operators to the sampled genome(s). This local interaction has its origin in constraints that derive from communication limitations in some distributed robotic scenarios. 
\cite{Schut2009a} showed it to be beneficial in simulated set-ups as an exploration / exploitation balancing mechanism. 

Embodied evolution, with chance encounters providing the sampling mechanism, has some similarities with  other flavours of evolutionary computation. Cellular evolutionary algorithms \citep{Alba2008Cellular-Geneti} consider continuous random rewiring of a network topology (in a grid of CPUs or computers) where all elements are evaluated in parallel. In this context, locally selecting candidates for reproduction is a recurring theme that is shared with embodied evolution (e.g. \cite{Garcia-Sanchez2012Testing-Diversi,FernandezPerez2014}).

\subsection{Objective Functions vs Selection Pressure}
\label{subsec:selectionpressure}

In traditional evolutionary algorithms, the optimisation process is guided by a (set of) objective function(s)~\citep{eiben2008book}. Evaluation of the candidate solutions, i.e., of the genomes in the population, allows for (typically numerical) comparison of their performance.
Beyond its relevance for performance assessment, the evaluation process \textit{per se} has generally no influence on the manner in which selection, variation and replacement evolutionary operators are applied. 
This is different in embodied evolution, where the behaviour of an individual can directly impact the likelihood of encounters with others and so influence selection and reproductive success~\citep{bredeche2010ppsn}.
Evolution can improve task performance, but it can also develop mating strategies, for example by maximising the number of encounters between robots if that improves the likelihood of transmitting genetic material. 

It is therefore important to realise that the \textit{selection pressure} on the robot population does not only derive from the specified \textit{objective function(s)} as it traditionally does in evolutionary computation. In embodied evolution, the environment, including the mechanisms that allow mating, also exert selection pressure. Consequently, evolution experiences selection pressure from the aggregate of objective function(s) and environmental particularities. 
\Citet{Steyven2016} researched how aspects of the robots' environment influence the emergence of particular behaviours and the balance between pressure towards survival and task.
The objective may even pose requirements that are opposed to those by the environment. This can be the case when a task implies risky behaviours, or because a task requires resources that are also needed for survival and mating.
In such situations, the evolutionary process must establish a trade-off between objective-driven optimisation and the maintenance of a viable environment where evolution occurs, which is a challenge in itself~\citep{Haasdijk2015SASO}. 

\subsection{Autonomous Performance Evaluation}
The decentralised nature of the evolutionary process implies that there is no omniscient presence who knows (let alone determines) the fitness values of all individuals.
Consequently, when an objective function is defined, it is the robots themselves that must  gauge their performance, and share it with other robots when mating: each robot must have an evaluation function that can be computed on-board and autonomously. 
Typical examples of such evaluation functions are the number of resources collected, the number of times a target has been reached or the number of collisions.
The requirement of autonomous assessment does not fundamentally change the way one defines fitness functions, but it does impact their usage as shown by
\citet{Nordin1997,Walker2006The-Balance-Bet,BredecheOn-line-On-boar} and \citet{Wolpert2008}.

\paragraph{Evaluation Time}
The robots must run a controller for some time to assess the resultant behaviour. This implies a {\em time sharing} scheme where robots run their current controllers to evaluate their performance. 
In many similar implementations, a robot runs a controller for a fixed evaluation time; \cite{haasdijk2012} showed that this is a very important parameter in encapsulated on\hyp{}line evolution, and it is likely to be similarly influential in embodied evolution. 

\paragraph{Evaluation in Varying Circumstances}
Because the evolutionary machinery (mating, evaluating new individuals, etc.) is an integral part of robot behaviour which runs in parallel with the performance of regular tasks, there can be no thorough re-initialisation or re-positioning procedure between genome replacements. This implies a noisy evaluation: a robot may undervalue a genome starting in adverse circumstances and vice versa.
As \citet[p.121]{Nordin1997} put it: \textit{``Each individual is tested against a different real-time situation leading to a unique fitness case. This results in `unfair' comparison where individuals have to navigate in situations with very different possible outcomes. However, [\ldots] experiments show that over time averaging tendencies of this learning method will even out the random effects of probabilistic sampling and a set of good solutions will survive''}.
\Citet{BredecheOn-line-On-boar} proposed a re-evaluation scheme to address this issue: seemingly efficient candidate solutions have a probability to be re-evaluated in order to cope with possible evaluation noise. A solution with a higher score \textit{and} a lower variance will then be preferred to one with a higher variance. While re-evaluation is not always used in embodied evolution, the evaluation of relatively similar genomes on different robots running in parallel provides another way to smooth the effect of noisy evaluations.

\paragraph{Multiple Objectives}
To deal with multiple objectives, evolutionary computation techniques typically selects individuals on the basis of Pareto dominance. While this is eminently possible when selecting partners as well, Pareto dominance can only be determined vis \`a vis the population sample that the selecting robot has acquired. It is unclear how this affects the overall performance and if the robot collective can effectively cover the Pareto front.
\citet{Haasdijk2014} and \citet{Bangel2017} investigated the use of a `market mechanism' to balance the selection pressure over multiple tasks in a concurrent foraging scenario, showing that this at least prevents the robot collective from focussing on single tasks, but that it does not lead to specialisation in individual robots.

\section{Discussion}
\label{sec:discussion}

The previous sections show that there is a considerable and increasing amount of research into embodied evolution, addressing issues that are particular to its autonomous and distributed nature. 
This section turns to the future of embodied evolution research, discussing potential applications and proposing a research agenda to tackle some of the more relevant and immediate issues that so far have remained insufficiently addressed in the field.

\subsection{Applications of embodied evolution}
\label{sec:applications}

Embodied evolution can be used as a design method for engineering, as a modelling method for evolutionary biology, or as a method to investigate evolving complex systems more generally. 
Let us briefly consider each of these possibilities.

\paragraph{Engineering} 
The on-line adaptivity afforded by embodied evolution offers many novel possibilities for deployment of robot collectives: exploration of unknown environments, search and rescue, distributed monitoring of large objects or areas, distributed construction, distributed mining, etc. 
Embodied evolution can offer a solution when robot collectives are required to be versatile, since the robots can be deployed in and adapt to open and a priori unknown environments and tasks. 
The collective is comparatively robust to failure through redundancy and the decentralised nature of the algorithm because the system continues to function even if some robots break down.
Embodied evolution can increase autonomy because the robots can, for instance, learn how to maintain energy while performing their task without intervention by an operator.

Currently, embodied evolution has already provided solutions to tasks such as navigation, surveillance and foraging  (see table \ref{tab:overview} for a complete list), but these are of limited interest because of the simplicity of the tasks considered in research to date. The research agenda proposed in section~\ref{sec:openissues} provides some suggestions for further research to mitigate this.

\paragraph{Evolutionary biology} 
In the last 100 years, evolutionary biology benefited from both experimental and theoretical advances. It is now possible, for instance, to study evolutionary mechanisms through methods such as gene sequencing~\citep{Blount2012,Wiser2013}. However, in vitro experimental evolution has its limitations: with evolution in ``real'' substrates, the time-scales involved limit the applicability to relatively simple organisms such as E.coli bacteria~\citep{Good2017}. 
From a theoretical point of view, population genetics (see \citet{Charlesworth2010} for a recent introduction) provides a set of mathematically grounded tools for understanding evolution dynamics, at the cost
of many simplifying assumptions.

Evolutionary robotics has recently gained relevance as an individual-based modelling and simulation method in evolutionary biology~\citep{Floreano2010,Waibel2011,long2012darwin,Mitri2013,Ferrante2015,Bernard2016pcb}, enabling the study of evolution in populations of robotic individuals in the physical world. 
Embodied evolution enables more accurate models of evolution because it is possible to embody not only the physical interactions, but also the evolutionary operators themselves.

\paragraph{Synthetic approach} 
Embodied evolution can also be used to ``understand by design'' \citep{Pfeifer2001}. As \citeauthor{maynard-smith1992}, a prominent researcher in evolutionary biology, advocated in a famous~\citeyear{maynard-smith1992}'s Science paper (originally referring to Tierra~\citep{Ray1993}): ``so far, we have been able to study only one evolving system and we cannot wait for interstellar flight to provide us with a second. If we want to discover generalisations about evolving systems, we have to look at artificial ones.'' 

This \textit{synthetic approach} stands somewhere between biology and engineering, using tools from the latter to understand mechanisms originally observed in nature, and aiming at identifying general principles not confined to any particular (biological) substrate. Beyond improving our \textit{understanding} of adaptive mechanisms, these general principles can also be used to improve our ability to \textit{design} complex systems.

\subsection{Research Agenda}
\label{sec:openissues}
We identify a number of open issues that need to be addressed so that embodied evolution can develop into a relevant technique to enable on-line adaptivity of robot collectives.
Some of these issues have been researched in other fields (e.g., credit assignment is a well-known and often considered topic in reinforcement learning research). Lessons can and should be learned from there, inspiring embodied evolution research into the relevance  and applicability of findings in those other fields.

In particular, we identify the following challenges:

\paragraph*{Benchmarks} 
The pseudo-code in section \ref{sec:algorithmic description} provides a clarification of embodied evolution's concepts by describing the basic building blocks of the algorithm.
This is only a first step towards a theoretical and practical framework for embodied evolution. Some authors have already taken steps in this direction. For instance, \citet{Prieto2015} propose an abstract algorithmic model in order to study both general and specific properties of embodied evolution implementations. \citet{Montanier2016} described  `vanilla' versions of embodied evolution algorithms that can be used as practical benchmarks. 
Further exploration of abstract models for theoretical validation is needed.
Also, standard benchmarks and test cases, along with systematically making the source code available, would provide a solid basis for empirical validation of individual contributions.

\paragraph*{Evolutionary Dynamics} 
Embodied evolution requires new tools for analysing the evolutionary dynamics at work. 
Because the evolutionary operators apply in situ, the dynamics of the  evolutionary process are not only important in the context of understanding or improving an optimisation procedure, but they also have a direct bearing on how the robots behave and change their behaviour when deployed.

Tools and methodologies to characterize the dynamics of evolving systems are available. The field of population genetics has produced techniques for estimating the selection pressure compared to genetic drift possibly occurring in finite-sized populations (see, for instance, \cite{Wakeley2008a} and \cite{Charlesworth2010} for a comprehensive introduction). Similarly, tools from adaptive dynamics~\citep{Geritz1998} can be used to investigate how particular solutions spread within the population. Finally, embodied evolution produces phylogenetic trees which can be studied either from a population genetics viewpoint (e.g. coalescence theory to understand the temporal structure of evolutionary adaptation) and graph theory (e.g. to characterize the particular structure of the inheritance graph). \citet{Boumaza2017} shows an interesting first foray into using this technique to analyse embodied evolution.

\paragraph*{Credit Assignment} 
In all the research reviewed in this paper that considers robot tasks, the fitness function is defined and implemented at the level of the individual robot: it assesses its own performance independently of the others. However, collectively solving a task often requires an assessment of performance at group level rather than individual level. This raises the issue of estimating each individual's contribution to the group's performance, which is unlikely to be completely captured by a fitness function (e.g. all individuals going towards the single larger food patch may not always be the best strategy if one aim to bring back the largest amount of food to the nest).

Closely related to our concern, \cite{stone2010} formulated the \textit{ad hoc teamwork} problem in multi-robot systems, involving robots that each must ``collaborate with previously unknown teammates on tasks to which they are all individually capable of contributing as team members''. As stated by \cite{Wolpert2008}, this implies devoting substantial attention to the problem of estimating the \textit{local utility} of individual agents with respect to the \textit{global welfare} of the whole group and how to make a trade-off between individual and group performance (e.g.,~\cite{Arthur1994,Hardin1968}).

While a generally applicable method to estimate an individual's local utility in an on-line distributed setting has so far eluded the community, it is possible to provide an exact assessment in controlled settings. Methods from cooperative game theory, such as computing the Shapley value~\citep{shapley1953}, could be used in embodied evolution but are computationally expensive and require the ability to replay experiments. However, replaying experiments is possible only with simulation and/or controlled experimental settings. While these methods cannot apply when robots are deployed in the real world, they at least provide a method to \textit{compare} the outcome of candidate solutions to estimate individuals' marginal contributions and choose which should be deployed.

\paragraph*{Social Complexity}
Section \ref{sec:state-of-the-art} shows that embodied evolution so far demonstrated only a limited set of social organisation concepts: simple cooperative and division of labour behaviours. In order to address more complex tasks, we must first gain a better understanding of the mechanisms required to achieve complex collective behaviours. This raises two questions. First, there is an \textit{ethological} question: what are the behavioural mechanisms at work in complex collective behaviours? Some of them, such as positive and negative feedback between individuals, or indirect communication through the environment (i.e., \textit{stigmergy}), are well known from examples found both in biology~\citep{Camazine2003a} and theoretical physics~\citep{Deutsch2012}. 
Secondly, there is a question about the origins and stability of behaviours: what are the key elements that make it possible to evolve collective behaviours, and what are their limits? Again, evolutionary ecology provides relevant insights, such as the interplay between the level of cooperation and relatedness between individuals \citep{West2007d}. 
The literature on such phenomena in biological systems may provide a good basis for research into the evolution of social complexity in embodied evolution.

A first step would be to clearly define the nature of social complexity that is to be studied. For this, evolutionary game theory~\citep{maynard-smith1992} has already produced a number of well-grounded and well-defined 'games` that capture many problems involving interactions among individuals, including thorough analysis of the evolutionary dynamics in simplified setups. Of course, results obtained on abstract models may not be transferable within more realistic settings (as \cite{Bernard2016pcb} showed for mutualistic cooperation), but the systematic use of a formal problem definition would greatly benefit the clarity of contributions in our domain. 

\paragraph*{Open-ended adaptation} 
As stated in Section \ref{sec:context}, embodied evolution aims to provide continuous adaptation so that the robot collective can cope with changes in the objectives and/or the environment. \citet{Montanier2011} showed that embodied evolution enables the population to react appropriately to changes in the regrowth rate of resources, but generally this aspect of embodied evolution has to date not been sufficiently addressed.

We reformulate the goal of continuous adaptation as providing \textit{open-ended} adaptation, i.e., having the ability to continually keep exploring new behavioural patterns, constructing increasingly complex behaviours as required. \cite{bedau2000}, \cite{soros2014}, \cite{Taylor2016} and others identified open-ended adaptation in artificial evolutionary systems as one of the big questions of artificial life.
Open-ended adaptation in artificial systems, in particular in combination with learning relevant task behaviour, has proved to be an elusive ambition.

A possible avenue to achieve this ambition may lie in the use of quality diversity approaches in embodied evolution.
Recent research has considered {\em quality diversity} measures as a replacement \citep{Lehman2011a} or additional \citep{mouret2012} objective to improve the population diversity and consequently the efficacy of evolution. 
To date, such research has focussed on the evolution of behaviour for particular tasks with task-specific metrics of behavioural diversity that must be tailored for each application.
To be able to exploit quality diversity in unknown environments and for arbitrary tasks generic measures of behavioural diversity must be developed.

Another avenue of research would be to take inspiration from the behaviour of a passerine bird, the great tit (parus major), as recently analysed by \cite{Aplin2017}. It appears that great tits combine collective and individual learning with varying intensity as they age, and that the motivations to pursue behaviours also vary with age. Reward-based learning occurs primarily in young birds and is often individual, while adult birds engage mostly in social learning to copy the behaviour that is most common, regardless of whether it produces more or less rewards than alternative behaviour. This combination of conformist and payoff-sensitive reinforcement allows individuals and populations both to acquire adaptive behaviour and to track environmental change. 
Combining embodied evolution, individual reinforcement learning with task-based and diversity-enhancing objectives may yield similar behavioural plasticity for collectives of robots.

\paragraph*{Safety and robot ethics}
To deploy the kind of adaptive technology that embodied evolution aims for responsibly, one must ensure that the adaptivity can be controlled: autonomous adaptation carries the risk of adaptation developing in directions that do not meet the needs of human users or that they even may find undesirable. Even so, the adaptive process should be curtailed as little as possible to allow effective, open-ended, learning. 
The user cannot be expected to monitor and closely control the robot’s behaviour and learning process; this may in fact be impossible in exactly those scenarios where robotic autonomy is most beneficial and adaptivity most urgently required. 
There is growing awareness that it may be necessary to endow robots with innately ethical behaviour \citep[e.g.,][]{Vanderelst2017,Anderson2007,Moor2006}, where the systems select actions based on a ``moral arithmetic'' \citep{Bentham1781}, often informed by casuistry, i.e., generalising morality on the basis of example cases in which there is agreement concerning the correct response \citep{Anderson2007}.
Moral reasoning along these lines could conceivably be enabled in embodied evolution as well, in which case interactive evolution to develop surrogate models of user requirements may offer one possible route to allow user guidance.

\paragraph*{}Additional open issues and opportunities will no doubt arise from advances in this and other fields. 
A relevant recent development, for instance, is the possibility of evolvable morpho-functional machines that are able to change both their software \textit{and} hardware features~\citep{Eiben2015} and replicate through 3D-printing~\citep{Brodbeck2015}. This would allow embodied evolution holistically to adapt the robots' morphologies as well as their controllers. This can have profound consequences for embodied evolution implementations that exploit these developments: it would, for instance, enable dynamic population sizes, allowing for more risky behaviour as broken robots could be replaced or recycled.

\section{Conclusions}

This paper provides an overview of embodied evolution for robot collectives, a research field that has been growing since its inception around the turn of the millennium. The main contribution of this paper is three-fold. First, it clarifies the definitions and overall process of embodied evolution. Second, it presents an overview of embodied evolution research conducted to date. Third, it provides directions for future researches.

This overview sheds light on the maturity of the field: while embodied evolution was mostly used as a parallel search method for designing individual behaviour during its first decade of existence, a trend has emerged towards its collective aspects (i.e., cooperation, division of labour, specialisation). This trend goes hand in hand with a trend towards larger, swarm-like, robot collectives.

We hope this overview will provide a stepping stone for the field, accounting for its maturity and acting as an inspiration for aspiring researchers. To this end, we  highlighted possible applications as well as open issues that may drive the field's research agenda.


\section*{Acknowledgements}
The authors gratefully acknowledge the support from the European Union’s Horizon 2020 research and innovation programme under grant agreement No 640891. The authors would also like to thank A.E. Eiben and Jean-Marc Montanier for their support during the writing of this paper.
\small

\bibliographystyle{apalike} 
\bibliography{EEreview}

\begin{thebibliography}{}

\bibitem[Alba, 2002]{Alba2002}
Alba, E. (2002).
\newblock {Parallel evolutionary algorithms can achieve super-linear
  performance}.
\newblock {\em Inf. Process. Lett.}, 82:7--13.

\bibitem[Alba and Dorronsoro, 2008]{Alba2008Cellular-Geneti}
Alba, E. and Dorronsoro, B. (2008).
\newblock {\em {Cellular Genetic Algorithms}}.
\newblock Springer--Verlag, Berlin--Heidelberg--New York.

\bibitem[Amato et~al., 2015]{amato2015}
Amato, C., Konidaris, G.~G., Cruz, G., Maynor, C.~A., How, J.~P., and
  Kaelbling, L.~P. (2015).
\newblock {Planning for decentralized control of multiple robots under
  uncertainty}.
\newblock In {\em 2015 IEEE International Conference on Robotics and Automation
  (ICRA)}, pages 1241--1248. IEEE.

\bibitem[Anderson and Anderson, 2007]{Anderson2007}
Anderson, M. and Anderson, S.~L. (2007).
\newblock {Machine ethics: Creating an ethical intelligent agent}.
\newblock {\em AI Mag.}, 28(4):15--26.

\bibitem[Aplin et~al., 2017]{Aplin2017}
Aplin, L.~M., Sheldon, B.~C., and McElreath, R. (2017).
\newblock {Conformity does not perpetuate suboptimal traditions in a wild
  population of songbirds}.
\newblock {\em Proc. Natl. Acad. Sci.}, 114(30):7830--7837.

\bibitem[Arthur, 1994]{Arthur1994}
Arthur, W. (1994).
\newblock {Inductive reasoning and bounded rationality}.
\newblock {\em The American economic review}, 84(2):406--411.

\bibitem[Axelrod, 1984]{axelrod1984the-evolution-o}
Axelrod, R. (1984).
\newblock {\em {The Evolution of Cooperation}}.
\newblock Basic Books, New York.

\bibitem[Bangel and Haasdijk, 2017]{Bangel2017}
Bangel, S. and Haasdijk, E. (2017).
\newblock {Reweighting rewards in embodied evolution to achieve a balanced
  distribution of labour}.
\newblock In {\em Proceedings of the 14th European Conference on Artificial
  Life ECAL 2017}, pages 44--51, Cambridge, MA. MIT Press.

\bibitem[Barrett et~al., 2016]{Barrett2016}
Barrett, S., Rosenfeld, A., Kraus, S., and Stone, P. (2016).
\newblock {Making friends on the fly: Cooperating with new teammates}.
\newblock {\em Artificial Intelligence}, pages 1--68.

\bibitem[Bayindir, 2016]{Bayindir2016}
Bayindir, L. (2016).
\newblock {A review of swarm robotics tasks}.
\newblock {\em Neurocomputing}, 172:292--321.

\bibitem[Bedau et~al., 2000]{bedau2000}
Bedau, M.~a., McCaskill, J.~S., Packard, N.~H., Rasmussen, S., Adami, C.,
  Green, D.~G., Ikegami, T., Kaneko, K., and Ray, T.~S. (2000).
\newblock {Open problems in artificial life.}
\newblock {\em Artificial life}, 6(4):363--76.

\bibitem[Bellingham and Rajan, 2007]{Bellingham2007a}
Bellingham, J.~G. and Rajan, K. (2007).
\newblock {Robotics in remote and hostile environments.}
\newblock {\em Science (New York, N.Y.)}, 318(5853):1098--1102.

\bibitem[Beni, 2005]{Beni2005}
Beni, G. (2005).
\newblock {From Swarm Intelligence to Swarm Robotics}.
\newblock {\em Robotics}, 3342:1--9.

\bibitem[Bentham, 1878]{Bentham1781}
Bentham, J. (1878).
\newblock {\em {Introduction to the Principles of Morals and Legislation}}.
\newblock Clarendon, Oxford, UK.

\bibitem[Bernard et~al., 2016]{Bernard2016pcb}
Bernard, A., Andr{\'{e}}, J.-B., and Bredeche, N. (2016).
\newblock {To Cooperate or Not to Cooperate: Why Behavioural Mechanisms
  Matter}.
\newblock {\em PLOS Computational Biology}, 12(5):e1004886.

\bibitem[Bernstein et~al., 2002]{Bernstein2002}
Bernstein, D. S.~., Givan, R., Immerman, N., and Zilberstein, S. (2002).
\newblock {The Complexity of Decentralized Control of Markov Decision
  Processes}.
\newblock {\em Mathematics of Operational Research}, 27(4):819--840.

\bibitem[Bianco and Nolfi, 2004]{Bianco2004}
Bianco, R. and Nolfi, S. (2004).
\newblock {Toward open-ended evolutionary robotics: evolving elementary robotic
  units able to self-assemble and self-reproduce}.
\newblock {\em Connection Science}, 16(4):227--248.

\bibitem[Blount et~al., 2012]{Blount2012}
Blount, Z.~D., Barrick, J.~E., Davidson, C.~J., and Lenski, R.~E. (2012).
\newblock {Genomic analysis of a key innovation in an experimental Escherichia
  coli population.}
\newblock {\em Nature}, 489(7417):513--8.

\bibitem[Bongard et~al., 2006]{Bongard2006b}
Bongard, J., Zykov, V., and Lipson, H. (2006).
\newblock {Resilient machines through continuous self-modeling.}
\newblock {\em Science (New York, N.Y.)}, 314(5802):1118--21.

\bibitem[Bongard, 2013]{Bongard2013}
Bongard, J.~C. (2013).
\newblock {Evolutionary Robotics}.
\newblock {\em Communications of the ACM}, 56(08):74--83.

\bibitem[Boumaza, 2017]{Boumaza2017}
Boumaza, A. (2017).
\newblock {Phylogeny of embodied evolutionary robotics}.
\newblock In {\em Proceedings of the Genetic and Evolutionary Computation
  Conference Companion on - GECCO '17}, pages 1681--1682, New York, New York,
  USA. ACM Press.

\bibitem[Brambilla et~al., 2012]{Brambilla2012}
Brambilla, M., Ferrante, E., Birattari, M., and Dorigo, M. (2012).
\newblock {Swarm robotics : A review from the swarm engineering perspective}.
\newblock {\em Swarm Intelligence}, 7(1):1--41.

\bibitem[Bredeche, 2014]{bredeche2014alife}
Bredeche, N. (2014).
\newblock {Embodied Evolutionary Robotics with Large Number of Robots}.
\newblock {\em Artificial Life 14: Proceedings of the Fourteenth International
  Conference on the Synthesis and Simulation of Living Systems}, pages
  272--273.

\bibitem[Bredeche et~al., 2009]{BredecheOn-line-On-boar}
Bredeche, N., Haasdijk, E., and Eiben, A.~E. (2009).
\newblock {On-line, On-board Evolution of Robot Controllers}.
\newblock In {\em Lecture Notes in Computer Science}, volume 5975, pages
  110--121. Springer Verlag.

\bibitem[Bredeche and Montanier, 2010]{bredeche2010ppsn}
Bredeche, N. and Montanier, J.-m. (2010).
\newblock {Environment-driven Embodied Evolution in a Population of Autonomous
  Agents}.
\newblock In {\em Parallel Problem Solving from Nature (PPSN)}, pages 290--299.

\bibitem[Bredeche and Montanier, 2012]{Bredeche2012}
Bredeche, N. and Montanier, J.-m. (2012).
\newblock {Environment-driven Open-ended Evolution with a Population of
  Autonomous Robots}.
\newblock In {\em Evolving Physical Systems Workshop}, pages 7--14.

\bibitem[Bredeche et~al., 2017]{Bredeche2017}
Bredeche, N., Montanier, J.-M., and Carrignon, S. (2017).
\newblock {Benefits of Proportionate Selection in Embodied Evolution : a Case
  Study with Behavioural Specialization}.
\newblock In {\em Proceedings of the Genetic and Evolutionary Computation
  Conference Companion}, pages 1683--1684.

\bibitem[Brodbeck et~al., 2015]{Brodbeck2015}
Brodbeck, L., Hauser, S., and Iida, F. (2015).
\newblock {Morphological Evolution of Physical Robots through Model-Free
  Phenotype Development.}
\newblock {\em PloS one}, 10(6):e0128444.

\bibitem[Camazine et~al., 2003]{Camazine2003a}
Camazine, S., Deneubourg, J.-L., Franks, N., Sneyd, J., Theraulaz, G., and
  Bonabeau, E. (2003).
\newblock {\em {Self-organization in biological systems}}.
\newblock Princeton University Press.

\bibitem[Charlesworth and Charlesworth, 2010]{Charlesworth2010}
Charlesworth, B. and Charlesworth, D. (2010).
\newblock {\em {Elements of evolutionary genetics}}.
\newblock Roberts and Company Publishers, Greenwood Village: Roberts and
  Company Publishers.

\bibitem[Christensen et~al., 2010]{Christensen2010}
Christensen, D.~J., Spr{\"{o}}witz, A., and Ijspeert, A.~J. (2010).
\newblock {Distributed Online Learning of Central Pattern Generators in Modular
  Robots}.
\newblock In {\em From Animals to Animats 11: Proceedings of the 11th
  international conference on Simulation of Adaptive Behavior}, pages 402--412.

\bibitem[Cully et~al., 2014]{Cully2014b}
Cully, A., Clune, J., and Mouret, J. (2014).
\newblock {Robots that can adapt like natural animals}.
\newblock {\em arXiv preprint arXiv:1407.3501}.

\bibitem[Deutsch et~al., 2012]{Deutsch2012}
Deutsch, A., Theraulaz, G., and Vicsek, T. (2012).
\newblock {Collective motion in biological systems}.
\newblock {\em Interface Focus}, 2(6):689--692.

\bibitem[Dibangoye et~al., 2015]{Dibangoye2014}
Dibangoye, J.~S., Amato, C., Buffet, O., and Charpillet, F. (2015).
\newblock {Exploiting separability in multiagent planning with continuous-state
  MDPs}.
\newblock In {\em IJCAI International Joint Conference on Artificial
  Intelligence}, volume 2015-Janua, pages 4254--4260.

\bibitem[Doncieux et~al., 2015]{Doncieux2015}
Doncieux, S., Bredeche, N., Mouret, J.-B., and Eiben, A. (2015).
\newblock {Evolutionary Robotics: What, Why, and Where to}.
\newblock {\em Frontiers in Robotics and AI}, 2(March):1--18.

\bibitem[Eiben and Smith, 2008]{eiben2008book}
Eiben, A. and Smith, J. (2008).
\newblock {\em {Introduction to Evolutionary Computing}}.
\newblock Springer.

\bibitem[Eiben et~al., 2010]{Eiben2010Embodied-On-lin}
Eiben, A.~E., Haasdijk, E., and Bredeche, N. (2010).
\newblock {Embodied, On-line, On-board Evolution for Autonomous Robotics}.
\newblock In Levi, P. and Kernbach, S., editors, {\em Symbiotic Multi-Robot
  Organisms: Reliability, Adaptability, Evolution}, chapter 5.2, pages
  361--382. Springer.

\bibitem[Eiben et~al., 2007]{Eiben2007Exploring-selec}
Eiben, A.~E., Schoenauer, M., Laredo, J. L.~J., Castillo, P.~A., Mora, A.~M.,
  and Merelo, J.~J. (2007).
\newblock {Exploring selection mechanisms for an agent-based distributed
  evolutionary algorithm}.
\newblock In {\em Proceedings of the 9th Annual Conference Companion on Genetic
  and Evolutionary Computation}, pages 2801--2808.

\bibitem[Eiben and Smith, 2015]{Eiben2015}
Eiben, A.~E. and Smith, J. (2015).
\newblock {From evolutionary computation to the evolution of things}.
\newblock {\em Nature}, 521(7553):476--482.

\bibitem[Fernandez~P{\'{e}}rez et~al., 2014]{FernandezPerez2014}
Fernandez~P{\'{e}}rez, I., Boumaza, A., and Charpillet, F. (2014).
\newblock {Comparison of Selection Methods in On-line Distributed Evolutionary
  Robotics}.
\newblock In {\em Proceedings of the fourteenth international conference on the
  synthesis and simulation of living systems}, pages 1--16.

\bibitem[Fernandez~P{\'{e}}rez et~al., 2015]{FernandezPerez2015}
Fernandez~P{\'{e}}rez, I., Boumaza, A., and Charpillet, F. (2015).
\newblock {Decentralized Innovation Marking for Neural Controllers in Embodied
  Evolution}.
\newblock In {\em Proceedings of the 2015 Annual Conference on Genetic and
  Evolutionary Computation}, pages 161--168.

\bibitem[Fernandez~P{\'{e}}rez et~al., 2017]{Perez2017}
Fernandez~P{\'{e}}rez, I., Boumaza, A., and Charpillet, F. (2017).
\newblock {Learning collaborative foraging in a swarm of robots using embodied
  evolution}.
\newblock In {\em Proceedings of the 14th European Conference on Artificial
  Life ECAL 2017}, pages 162--161, Cambridge, MA. MIT Press.

\bibitem[Ferrante et~al., 2015]{Ferrante2015}
Ferrante, E., Turgut, A.~E., Du{\'{e}}{\~{n}}ez-Guzman, E., Dorigo, M., and
  Wenseleers, T. (2015).
\newblock {Evolution of Self-Organized Task Specialization in Robot Swarms}.
\newblock {\em PLoS Computational Biology}, 11(8):e1004273.

\bibitem[Ficici et~al., 1999]{Ficici1999}
Ficici, S.~G., Watson, R.~A., and Pollack, J.~B. (1999).
\newblock {Embodied Evolution: A Response to Challenges in Evolutionary
  Robotics}.
\newblock In Wyatt, J.~L. and Demiris, J., editors, {\em Proceedings of the
  Eighth European Workshop on Learning Robots}, pages 14--22.

\bibitem[Floreano and Keller, 2010]{Floreano2010}
Floreano, D. and Keller, L. (2010).
\newblock {Evolution of adaptive behaviour in robots by means of Darwinian
  selection.}
\newblock {\em PLoS biology}, 8(1):1--8.

\bibitem[Garc{\'{i}}a-S{\'{a}}nchez et~al.,
  2012]{Garcia-Sanchez2012Testing-Diversi}
Garc{\'{i}}a-S{\'{a}}nchez, P., Eiben, a.~E., Haasdijk, E., Weel, B., and
  Merelo-Guerv{\'{o}}s, J.-J.~J. (2012).
\newblock {Testing Diversity-Enhancing Migration Policies for Hybrid On-Line
  Evolution of Robot Controllers}.
\newblock In {\em Lecture Notes in Computer Science (including subseries
  Lecture Notes in Artificial Intelligence and Lecture Notes in
  Bioinformatics)}, volume 7248 LNCS, pages 52--62.

\bibitem[Gauci et~al., 2012]{gauci2013evolving}
Gauci, M., Chen, J., Dodd, T.~J., and Gro{\ss}, R. (2012).
\newblock {Evolving Aggregation Behaviors in Multi-Robot Systems with Binary
  Sensors}.
\newblock In {\em 2012 Int. Symposium on Distributed Autonomous Robotic Systems
  (DARS 2012)}. Springer.

\bibitem[Geritz et~al., 1998]{Geritz1998}
Geritz, S., Kisdi, E., Meszena, G., and Metz, J. (1998).
\newblock {Evolutionarily singular strategies and the adaptive growth and
  branching of the evolutionary tree}.
\newblock {\em Evolutionary ecology}, 12:35--57.

\bibitem[Good et~al., 2017]{Good2017}
Good, B.~H., McDonald, M.~J., Barrick, J.~E., Lenski, R.~E., and Desai, M.~M.
  (2017).
\newblock {The dynamics of molecular evolution over 60,000 generations}.
\newblock {\em Nature}, 551(7678):45--50.

\bibitem[Haasdijk, 2015]{Haasdijk2015SASO}
Haasdijk, E. (2015).
\newblock {Combining Conflicting Environmental and Task Requirements in
  Evolutionary Robotics}.
\newblock In {\em 2015 IEEE 9th International Conference on Self-Adaptive and
  Self-Organizing Systems}, pages 131--137.

\bibitem[Haasdijk and Bredeche, 2013]{Haasdijk2013}
Haasdijk, E. and Bredeche, N. (2013).
\newblock {Controlling Task Distribution in MONEE}.
\newblock In {\em Advances in Artificial Life, ECAL 2013}, pages 671--678. MIT
  Press.

\bibitem[Haasdijk et~al., 2014a]{Haasdijk2014}
Haasdijk, E., Bredeche, N., and Eiben, a.~E. (2014a).
\newblock {Combining environment-driven adaptation and task-driven optimisation
  in evolutionary robotics.}
\newblock {\em PloS one}, 9(6):e98466.

\bibitem[Haasdijk et~al., 2014b]{Haasdijk2014a}
Haasdijk, E., Bredeche, N., Nolfi, S., and Eiben, A.~E. (2014b).
\newblock {Evolutionary robotics}.
\newblock {\em Evolutionary Intelligence}, 7(2):69--70.

\bibitem[Haasdijk and Eigenhuis, 2016]{Haasdijk2016}
Haasdijk, E. and Eigenhuis, F. (2016).
\newblock {Increasing Reward in Biased Natural Selection Decreases Task
  Performance}.
\newblock In {\em Artificial Life {\{}XV{\}} : Proceedings of the 15th
  International Conference on Artificial Life}, pages 314--322.

\bibitem[Haasdijk et~al., 2012]{haasdijk2012}
Haasdijk, E., Smit, S.~K., and Eiben, A.~E. (2012).
\newblock {Exploratory analysis of an on-line evolutionary algorithm in
  simulated robots}.
\newblock {\em Evolutionary Intelligence}, 5(4):213--230.

\bibitem[Haasdijk et~al., 2013]{Haasdijk2013b}
Haasdijk, E., Weel, B., and Eiben, A.~E. (2013).
\newblock {Right on the MONEE}.
\newblock In Blum, C., editor, {\em Proceeding of the fifteenth annual
  conference on Genetic and evolutionary computation conference}, pages
  207--214. ACM New York, NY, USA.

\bibitem[Hardin, 1968]{Hardin1968}
Hardin, G. (1968).
\newblock {The Tragedy of the Commons}.
\newblock {\em Science}, 162(June):1243--1248.

\bibitem[Hart et~al., 2015]{Hart2015}
Hart, E., Steyven, A., and Paechter, B. (2015).
\newblock {Improving Survivability in Environment-driven Distributed
  Evolutionary Algorithms through Explicit Relative Fitness and Fitness
  Proportionate Communication}.
\newblock In {\em Proceedings of the 2015 Annual Conference on Genetic and
  Evolutionary Computation}, pages 169--176.

\bibitem[Hauert et~al., 2008]{Hauert2008}
Hauert, S., Zufferey, J.-C., and Floreano, D. (2008).
\newblock {Evolved swarming without positioning information: an application in
  aerial communication relay}.
\newblock {\em Autonomous Robots}, 26(1):21--32.

\bibitem[Heinerman et~al., 2015]{Heinerman2015-3foldadaptivity}
Heinerman, J., Drupsteen, D., and Eiben, A.~E. (2015).
\newblock {Three-fold Adaptivity in Groups of Robots: The Effect of Social
  Learning}.
\newblock In Silva, S., editor, {\em Proceedings of the 17th annual conference
  on Genetic and evolutionary computation}, GECCO '15, pages 177--183. ACM.

\bibitem[Heinerman et~al., 2016]{Heinerman2015-3foldthymio}
Heinerman, J., Rango, M., and Eiben, A.~E. (2016).
\newblock {Evolution, individual learning, and social learning in a swarm of
  real robots}.
\newblock In {\em Proceedings - 2015 IEEE Symposium Series on Computational
  Intelligence, SSCI 2015}, pages 1055--1062. IEEE.

\bibitem[Hettiarachchi and Spears, 2009]{Hettiarachchi2009a}
Hettiarachchi, S. and Spears, W.~M. (2009).
\newblock {Distributed adaptive swarm for obstacle avoidance}.
\newblock {\em International Journal of Intelligent Computing and Cybernetics},
  2(4):644--671.

\bibitem[Hettiarachchi et~al., 2006]{Hettiarachchi2006}
Hettiarachchi, S., Spears, W.~M., Green, D., and Kerr, W. (2006).
\newblock {Distributed agent evolution with dynamic adaptation to local
  unexpected scenarios}.
\newblock In Hinchey, M.~G., Rago, P., Rash, J.~L., Rouff, C.~A., Sterritt, R.,
  and Truszkowski, W., editors, {\em Innovative Concepts for Autonomic and
  Agent-Based Systems}, volume LNCS 3825, pages 245--256, Greenbelt, MD.
  Springer.

\bibitem[Huijsman et~al., 2011]{Huijsman2011An-On-line-On-b}
Huijsman, R.-J., Haasdijk, E., and Eiben, A.~E. (2011).
\newblock {An On-line On-board Distributed Algorithm for Evolutionary
  Robotics}.
\newblock In Hao, J.-K., Legrand, P., Collet, P., Monmarch{\'{e}}, N., Lutton,
  E., and Schoenauer, M., editors, {\em Artificial Evolution, 10th
  International Conference Evolution Artificielle}, number 7401 in LNCS, pages
  73--84. Springer.

\bibitem[Jakobi et~al., 1995]{Jakobi1995}
Jakobi, N., Husbands, P., and Harvey, I. (1995).
\newblock {Noise and the Reality Gap: The Use of Simulation in Evolutionary
  Robotics}.
\newblock {\em Lecture Notes in Computer Science}, 929:704--720.

\bibitem[Karafotias et~al., 2011]{Karafotias2011An-Algorithm-fo}
Karafotias, G., Haasdijk, E., and Eiben, A.~E. (2011).
\newblock {An Algorithm for Distributed On-line, On-board Evolutionary
  Robotics}.
\newblock In {\em GECCO '11: Proceedings of the 13th annual conference on
  Genetic and evolutionary computation}, pages 171--178. ACM Press.

\bibitem[Kemeling and Haasdijk, 2017]{Kemeling2017}
Kemeling, M. and Haasdijk, E. (2017).
\newblock {Incorporating user feedback in embodied evolution}.
\newblock In {\em Proceedings of the Genetic and Evolutionary Computation
  Conference Companion on - GECCO '17}, pages 1685--1686, New York, New York,
  USA. ACM Press.

\bibitem[K{\"{o}}nig et~al., 2009]{Konig2009b}
K{\"{o}}nig, L., Mostaghim, S., and Schmeck, H. (2009).
\newblock {Decentralized evolution of Robotic behavior using Finite State
  Automata}.
\newblock {\em International Journal of Intelligent Computing and Cybernetics},
  2(4)(4):695--723.

\bibitem[K{\"{o}}nig and Schmeck, 2009]{Konig2009a}
K{\"{o}}nig, L. and Schmeck, H. (2009).
\newblock {A completely evolvable genotype-phenotype mapping for evolutionary
  robotics}.
\newblock In {\em SASO 2009 - 3rd IEEE International Conference on
  Self-Adaptive and Self-Organizing Systems}, pages 175--185.

\bibitem[Lehman and Stanley, 2011]{Lehman2011a}
Lehman, J. and Stanley, K.~O. (2011).
\newblock {Abandoning objectives: evolution through the search for novelty
  alone}.
\newblock {\em Evolutionary computation}, 19(2):189--223.

\bibitem[Long, 2012]{long2012darwin}
Long, J. (2012).
\newblock {\em {Darwin's Devices: What Evolving Robots Can Teach Us about the
  History of Life and the Future of Technology}}.
\newblock Basic Books.

\bibitem[Mataric, 1994]{Mataric1994}
Mataric, M.~J. (1994).
\newblock {\em {Interaction and Intelligent Behavior}}.
\newblock PhD thesis.

\bibitem[Maynard~Smith, 1992]{maynard-smith1992}
Maynard~Smith, J. (1992).
\newblock {Evolutionary biology. Byte-sized evolution.}
\newblock {\em Nature}, 355(6363):772--773.

\bibitem[Mitri et~al., 2013]{Mitri2013}
Mitri, S., Wischmann, S., Floreano, D., and Keller, L. (2013).
\newblock {Using robots to understand social behaviour.}
\newblock {\em Biological reviews of the Cambridge Philosophical Society},
  88(1):31--9.

\bibitem[Montanier and Bredeche, 2011]{Montanier2011}
Montanier, J.-m. and Bredeche, N. (2011).
\newblock {Surviving the Tragedy of Commons: Emergence of Altruism in a
  Population of Evolving Autonomous Agents}.
\newblock In {\em European Conference on Artificial Life}, pages 550--557,
  Paris, France.

\bibitem[Montanier and Bredeche, 2013]{montanier2013}
Montanier, J.-M. and Bredeche, N. (2013).
\newblock {Evolution of Altruism and Spatial Dispersion: an Artificial
  Evolutionary Ecology Approach}.
\newblock {\em Advances in Artificial Life, ECAL 2013}, pages 260--267.

\bibitem[Montanier et~al., 2016]{Montanier2016}
Montanier, J.-m., Carrignon, S., and Bredeche, N. (2016).
\newblock {Behavioural Specialization in Embodied Evolutionary Robotics: Why so
  Difficult ?}
\newblock {\em Frontiers in Robotics and AI}, pages 1--17.

\bibitem[Moor, 2006]{Moor2006}
Moor, J.~H. (2006).
\newblock {The nature, importance, and difficulty of machine ethics}.
\newblock {\em {\{}{\{}{\}}IEEE{\{}{\}}{\}} Intell. Syst.}, 21(4):18--21.

\bibitem[Mouret and Doncieux, 2012]{mouret2012}
Mouret, J.~B. and Doncieux, S. (2012).
\newblock {Encouraging Behavioral Diversity in Evolutionary Robotics: an
  Empirical Study}.
\newblock {\em Evol. Comput.}, 20(1):91--133.

\bibitem[Mouret and Tonelli, 2015]{Mouret2015}
Mouret, J.~B. and Tonelli, P. (2015).
\newblock {Artificial evolution of plastic neural networks: A few key
  concepts}.
\newblock {\em Studies in Computational Intelligence}, 557:251--261.

\bibitem[Nelson and Grant, 2006]{Nelson2006}
Nelson, A.~L. and Grant, E. (2006).
\newblock {Using direct competition to select for competent controllers in
  evolutionary robotics}.
\newblock {\em Robotics and Autonomous Systems}, 54(10):840--857.

\bibitem[Nolfi and Floreano, 2000]{Nolfi2000}
Nolfi, S. and Floreano, D. (2000).
\newblock {\em {Evolutionary Robotics: the Biology, Intelligence, and
  Technology}}.
\newblock MIT Press, Cambridge MA, USA.

\bibitem[Nordin and Banzhaf, 1997]{Nordin1997}
Nordin, P. and Banzhaf, W. (1997).
\newblock {An On-Line Method to Evolve Behavior and to Control a Miniature
  Robot in Real Time with Genetic Programming}.
\newblock {\em Adaptive Behavior}, 5(2):107--140.

\bibitem[Noskov et~al., 2013]{Noskov2013a}
Noskov, N., Haasdijk, E., Weel, B., and Eiben, a.~E. (2013).
\newblock {MONEE: Using parental investment to combine open-ended and
  task-driven evolution}.
\newblock {\em Lecture Notes in Computer Science (including subseries Lecture
  Notes in Artificial Intelligence and Lecture Notes in Bioinformatics)}, 7835
  LNCS:569--578.

\bibitem[Nouyan et~al., 2009]{Nouyan2009}
Nouyan, S., Gross, R., Bonani, M., Mondada, F., and Dorigo, M. (2009).
\newblock {Teamwork in Self-Organized Robot Colonies}.
\newblock {\em IEEE Transactions on Evolutionary Computation}, 13(4):695--711.

\bibitem[O'Dowd et~al., 2014]{ODowd2014a}
O'Dowd, P.~J., Studley, M., and Winfield, A. F.~T. (2014).
\newblock {The distributed co-evolution of an on-board simulator and controller
  for swarm robot behaviours}.
\newblock {\em Evolutionary Intelligence}, 7(2):95--106.

\bibitem[Parker, 2008]{Parker2008a}
Parker, L.~E. (2008).
\newblock {Multiple Mobile Robot Systems}.
\newblock In Heidelberg, S.~B., editor, {\em Handbook of Robotics}, chapter~40,
  pages 921--941. Springer.

\bibitem[Perez et~al., 2008]{Perez2008Embodied-Evolut}
Perez, A. L.~F., Bittencourt, G., and Roisenberg, M. (2008).
\newblock {Embodied evolution with a new genetic programming variation
  algorithm}.
\newblock In {\em Proceedings - 4th International Conference on Autonomic and
  Autonomous Systems, ICAS 2008}, volume~0, pages 118--123, Los Alamitos, CA,
  USA. IEEE Press.

\bibitem[Pfeifer and Scheier, 2001]{Pfeifer2001}
Pfeifer, R. and Scheier, C. (2001).
\newblock {\em {Understanding Intelligence}}.
\newblock MIT Press.

\bibitem[Prieto et~al., 2010]{Prieto2010}
Prieto, A., Becerra, J., Bellas, F., and Duro, R. (2010).
\newblock {Open-ended evolution as a means to self-organize heterogeneous
  multi-robot systems in real time}.
\newblock {\em Robotics and Autonomous Systems}, 58(12):1282--1291.

\bibitem[Prieto et~al., 2009]{Prieto2009}
Prieto, A., Bellas, F., and Duro, R.~J. (2009).
\newblock {Adaptively coordinating heterogeneous robot teams through
  asynchronous situated coevolution}.
\newblock {\em Lecture Notes in Computer Science}, 5864 LNCS(PART 2):75--82.

\bibitem[Prieto et~al., 2015]{Prieto2015}
Prieto, A., Bellas, F., Trueba, P., and Duro, R.~J. (2015).
\newblock {Towards the standardization of distributed Embodied Evolution}.
\newblock {\em Information Sciences}, 312:55--77.

\bibitem[Prieto et~al., 2016]{Prieto2016}
Prieto, A., Bellas, F., Trueba, P., and Duro, R.~J. (2016).
\newblock {Real-time optimization of dynamic problems through distributed
  Embodied Evolution}.
\newblock {\em Integrated Computer-Aided Engineering}, 23(3):237--253.

\bibitem[Pugh and Martinoli, 2009]{Pugh2009}
Pugh, J. and Martinoli, A. (2009).
\newblock {Distributed scalable multi-robot learning using particle swarm
  optimization}.
\newblock {\em Swarm Intelligence}, 3(3):203--222.

\bibitem[Ray, 1993]{Ray1993}
Ray, T.~S. (1993).
\newblock {An Evolutionary Approach to Synthetic Biology: Zen and the Art of
  Creating Life}.
\newblock {\em Artificial Life}, 1:179--209.

\bibitem[Rubenstein et~al., 2014]{Rubenstein2014}
Rubenstein, M., Cornejo, a., and Nagpal, R. (2014).
\newblock {Programmable self-assembly in a thousand-robot swarm}.
\newblock {\em Science}, 345(6198):795--799.

\bibitem[Schut et~al., 2009]{Schut2009a}
Schut, M.~C., Haasdijk, E., and Prieto, A. (2009).
\newblock {Is situated evolution an alternative for classical evolution?}
\newblock In {\em 2009 IEEE Congress on Evolutionary Computation, CEC 2009},
  pages 2971--2976.

\bibitem[Schwarzer et~al., 2010]{Schwarzer2010Artificial-Sexu}
Schwarzer, C., H{\"{o}}sler, C., and Michiels, N. (2010).
\newblock {Artificial Sexuality and Reproduction of Robot Organisms}.
\newblock In Levi, P. and Kernbach, S., editors, {\em Symbiotic Multi-Robot
  Organisms: Reliability, Adaptability, Evolution}, pages 384--403.
  Springer--Verlag, Berlin--Heidelberg--New York.

\bibitem[Schwarzer et~al., 2011]{Schwarzer2011}
Schwarzer, C., Schlachter, F., and Michiels, N.~K. (2011).
\newblock {Online Evolution in Dynamic Environments using Neural Networks in
  Autonomous Robots}.
\newblock {\em International Journal On Advances in Intelligent Systems},
  4(3--4):288--298.

\bibitem[Shapley, 1953]{shapley1953}
Shapley, L.~S. (1953).
\newblock {A Value for n-person Games}.
\newblock {\em Annals of Mathematical Studies}, 28:307–317.

\bibitem[Silva et~al., 2013]{Silva2013a}
Silva, F., Correia, L., and Christensen, A.~L. (2013).
\newblock {Dynamics of neuronal models in online neuroevolution of robotic
  controllers}.
\newblock {\em Lecture Notes in Computer Science (including subseries Lecture
  Notes in Artificial Intelligence and Lecture Notes in Bioinformatics)}, 8154
  LNAI:90--101.

\bibitem[Silva et~al., 2017]{Silva2017}
Silva, F., Correia, L., and Christensen, A.~L. (2017).
\newblock {Evolutionary online behaviour learning and adaptation in real
  robots}.
\newblock {\em Royal Society Open Science}, 4(7):160938.

\bibitem[Silva et~al., 2016]{Silva2016}
Silva, F., Duarte, M., Correia, L., Oliveira, S.~M., and Christensen, A.~L.
  (2016).
\newblock {Open Issues in Evolutionary Robotics}.
\newblock {\em Evolutionary Computation}, 24(2):205--236.

\bibitem[Silva et~al., 2015]{Silva2015}
Silva, F., Urbano, P., Correia, L., and Christensen, A.~L. (2015).
\newblock {odNEAT: An Algorithm for Decentralised Online Evolution of Robotic
  Controllers}.
\newblock {\em Evolutionary Computation}, 23(3):421--449.

\bibitem[Silva et~al., 2012]{Silva2012}
Silva, F., Urbano, P., Oliveira, S., and Christensen, A.~L. (2012).
\newblock {odNEAT: An Algorithm for Distributed Online, Onboard Evolution of
  Robot Behaviours}.
\newblock {\em Artificial Life 13}, pages 251--258.

\bibitem[Simões and Dimond, 2001]{Simoes2001}
Simões, E. d.~V. and Dimond, K. R.~K. (2001).
\newblock {Embedding a distributed evolutionary system into a population of
  autonomous mobile robots}.
\newblock {\em 2001 IEEE International Conference on Systems, Man and
  Cybernetics}, 2:1069--1074.

\bibitem[Soros and Stanley, 2014]{soros2014}
Soros, L.~B. and Stanley, K. K.~O. (2014).
\newblock {Identifying Necessary Conditions for Open-Ended Evolution through
  the Artificial Life World of Chromaria}.
\newblock In {\em Proc. of Artificial Life Conference (ALife 14)}, pages
  793--800.

\bibitem[Stanley et~al., 2005]{Stanley2005}
Stanley, K.~O., Bryant, B.~D., and Miikkulainen, R. (2005).
\newblock {Real-time neuroevolution in the NERO video game}.
\newblock {\em IEEE Transactions on Evolutionary Computation}, 9(6):653--668.

\bibitem[Steyven et~al., 2016]{Steyven2016}
Steyven, A., Hart, E., and Paechter, B. (2016).
\newblock {Understanding Environmental Influence in an Open-Ended Evolutionary
  Algorithm}.
\newblock In Handl, J., Hart, E., Lewis, P.~R., L{\'{o}}pez-Ib{\'{a}}{\~{n}}ez,
  M., Ochoa, G., and Paechter, B., editors, {\em Parallel Problem Solving from
  Nature -- PPSN XIV: 14th International Conference, Edinburgh, UK, September
  17-21, 2016, Proceedings}, pages 921--931, Cham. Springer International
  Publishing.

\bibitem[Stone et~al., 2010]{stone2010}
Stone, P., Kaminka, G.~A., Kraus, S., and Rosenschein, J.~S. (2010).
\newblock {Ad Hoc Autonomous Agent Teams : Collaboration without
  Pre-Coordination}.
\newblock In {\em The Twenty-Fourth Conference on Artificial Intelligence
  (AAAI)}, number July.

\bibitem[Stone et~al., 2005]{Stone2005a}
Stone, P., Sutton, R.~S., and Kuhlmann, G. (2005).
\newblock {Reinforcement Learning for RoboCup-Soccer Keepaway}.
\newblock {\em Adaptive Behavior}, 13(3):165--188.

\bibitem[Stone and Veloso, 1998]{Stone1998}
Stone, P. and Veloso, M. (1998).
\newblock {Layered approach to learning client behaviors in the robocup soccer
  server}.
\newblock {\em Applied Artificial Intelligence}, 12(2-3):165--188.

\bibitem[Taylor et~al., 2016]{Taylor2016}
Taylor, T., Bedau, M., Channon, A., Ackley, D., Banzhaf, W., Beslon, G.,
  Dolson, E., Froese, T., Hickinbotham, S., Ikegami, T., McMullin, B., Packard,
  N., Rasmussen, S., Virgo, N., Agmon, E., Clark, A., McGregor, S., Ofria, C.,
  Ropella, G., Spector, L., Stanley, K.~O., Stanton, A., Timperley, C.,
  Vostinar, A., and Wiser, M. (2016).
\newblock {Open-Ended Evolution : Perspectives from the OEE Workshop in York}.
\newblock {\em Artificial Life}, 22:408--423.

\bibitem[Thrun and Mitchell, 1995]{Thrun1995}
Thrun, S. and Mitchell, T.~M. (1995).
\newblock {Lifelong robot learning}.
\newblock {\em Robotics and Autonomous Systems}, 15:25--46.

\bibitem[Tonelli and Mouret, 2013]{Tonelli2013c}
Tonelli, P. and Mouret, J.~B. (2013).
\newblock {On the relationships between generative encodings, regularity, and
  learning abilities when evolving plastic artificial neural networks}.
\newblock {\em PLoS ONE}, 8(11).

\bibitem[Trianni et~al., 2008]{Trianni2008}
Trianni, V., Nolfi, S., and Dorigo, M. (2008).
\newblock {Evolution, Self-organization and Swarm Robotics}.
\newblock In {\em Swarm Intelligence}, pages 163--191.

\bibitem[Trueba, 2017]{Trueba2017}
Trueba, P. (2017).
\newblock {Embodied Evolution versus Cooperative Coevolution in Multi- Robot
  Optimization : a practical comparison}.
\newblock pages 79--80.

\bibitem[Trueba et~al., 2011]{Trueba2011}
Trueba, P., Prieto, A., Bellas, F., Caama{\~{n}}o, P., and Duro, R.~J. (2011).
\newblock {Task-driven species in evolutionary robotic teams}.
\newblock In {\em International Work-Conference on the Interplay Between
  Natural and Artificial Computation}, pages 138--147. Springer.

\bibitem[Trueba et~al., 2012]{Trueba2012a}
Trueba, P., Prieto, A., Bellas, F., Caama{\~{n}}o, P., and Duro, R.~J. (2012).
\newblock {Self-organization and specialization in multiagent systems through
  open-ended natural evolution}.
\newblock In {\em Lecture Notes in Computer Science}, volume 7248 LNCS of {\em
  Lecture Notes in Computer Science}, pages 93--102. Springer Berlin
  Heidelberg.

\bibitem[Trueba et~al., 2013]{Trueba2013682}
Trueba, P., Prieto, A., Bellas, F., Caama{\~{n}}o, P., and Duro, R.~J. (2013).
\newblock {Specialization analysis of embodied evolution for robotic collective
  tasks}.
\newblock {\em Robotics and Autonomous Systems}, 61(7):682--693.

\bibitem[Urzelai and Floreano, 2001]{Urzelai2001b}
Urzelai, J. and Floreano, D. (2001).
\newblock {Evolution of adaptive synapses: robots with fast adaptive behavior
  in new environments.}
\newblock {\em Evolutionary computation}, 9(4):495--524.

\bibitem[Usui and Arita, 2003]{Usui2003Situated-and-Em}
Usui, Y. and Arita, T. (2003).
\newblock {Situated and Embodied Evolution in Collective Evolutionary
  Robotics}.
\newblock In {\em Proceedings of the 8th International Symposium on Artificial
  Life and Robotics}, number~c, pages 212--215.

\bibitem[Vanderelst and Winfield, 2017]{Vanderelst2017}
Vanderelst, D. and Winfield, A. (2017).
\newblock {An architecture for ethical robots inspired by the simulation theory
  of cognition}.
\newblock {\em Cogn. Syst. Res.}

\bibitem[Waibel et~al., 2011]{Waibel2011}
Waibel, M., Floreano, D., and Keller, L. (2011).
\newblock {A quantitative test of Hamilton's rule for the evolution of
  altruism}.
\newblock {\em PLoS biology}, 9(5):1--7.

\bibitem[Wakeley, 2008]{Wakeley2008a}
Wakeley, J. (2008).
\newblock {\em {Coalescent theory, an introduction}}.
\newblock Roberts {\&} Company Publishers.

\bibitem[Walker et~al., 2006]{Walker2006The-Balance-Bet}
Walker, J.~H., Garrett, S.~M., and Wilson, M.~S. (2006).
\newblock {The Balance Between Initial Training and Lifelong Adaptation in
  Evolving Robot Controllers}.
\newblock {\em {\{}IEEE{\}} Trans. Syst., Man, Cybern. {\{}B{\}}},
  36(2):423--432.

\bibitem[Watson et~al., 2002]{Watson2002a}
Watson, R.~A., Ficici, S.~G., and Pollack, J.~B. (2002).
\newblock {Embodied Evolution: Distributing an evolutionary algorithm in a
  population of robots}.
\newblock {\em Robotics and Autonomous Systems}, 39(1):1--18.

\bibitem[Weel et~al., 2012a]{weel2012organisms}
Weel, B., Haasdijk, E., and Eiben, A.~E. (2012a).
\newblock {The Emergence of Multi-Cellular Robot Organisms through On-line
  On-board Evolution}.
\newblock In Di~Chio {\textbackslash}em~et al, C., editor, {\em Applications of
  Evolutionary Computation}, volume 7248 of {\em Lecture Notes in Computer
  Science}, pages 124--134. Springer.

\bibitem[Weel et~al., 2012b]{weel2012online}
Weel, B., Hoogendoorn, M., and Eiben, A.~E. (2012b).
\newblock {On-Line Evolution of Controllers for Aggregating Swarm Robots in
  Changing Environments}.
\newblock In {\em PPSN XII: Proceedings of the 12th International Conference on
  Parallel Problem Solving from Nature, LNCS 7492}, pages 245--254.

\bibitem[Werfel et~al., 2014]{Werfel2014a}
Werfel, J., Petersen, K., and Nagpal, R. (2014).
\newblock {Designing collective behavior in a Termite-Inspired robot
  construction team}.
\newblock {\em Science}, 754.

\bibitem[West et~al., 2007]{West2007d}
West, S.~A., Griffin, A.~S., and Gardner, A. (2007).
\newblock {Social semantics: altruism, cooperation, mutualism, strong
  reciprocity and group selection.}
\newblock {\em Journal of evolutionary biology}, 20(2):415--32.

\bibitem[Wischmann et~al., 2007]{Wischmann2007Embodied-evolut}
Wischmann, S., Stamm, K., and Florentin, W. (2007).
\newblock {Embodied Evolution and Learning : The Neglected Timing of
  Maturation}.
\newblock In {\em ECAL 2007: Advances in Artificial Life}, volume 4648, pages
  284--293. Springer.

\bibitem[Wiser et~al., 2013]{Wiser2013}
Wiser, M.~J., Ribeck, N., and Lenski, R.~E. (2013).
\newblock {Long-term dynamics of adaptation in asexual populations.}
\newblock {\em Science}, 342(6164):1364--7.

\bibitem[Wolpert and Tumer, 2008]{Wolpert2008}
Wolpert, D.~H. and Tumer, K. (2008).
\newblock {An introduction to collective intelligence}.
\newblock Technical report, NASA.

\end{thebibliography}

\end{document}